**Title**

Arranging Test Tubes in Racks Using Combined Task and Motion Planning

**Authors**

W. Wan[1*], T. Kotaka[2], K. Harada[1]

**Affiliations**

[1]Graduate School of Engineering Science, Osaka University, Japan.
[2]Miraca Research Institute G. K., Japan.

*Correspondent author

**Abstract**

A robot manipulation system that separates and arranges test tubes in racks with the help of 3D vision and artificial intelligence (AI) reasoning/planning.

## MAIN TEXT

The large amount of infections by COVID-19 drives people to perform thousands of polymerase chain reaction tests, antibody tests, etc. These tests require to handle a huge amount of test tubes, which are not only labor-intensive but also pressing. Employing a simple-to-use robot to do the job and consequently replace human labor is highly expected.

In Japan, we are collaborating with Miraca Research Institute G.K. (MRI) to develop a robotic manipulation system that helps to handle the test tubes for clinical examinations. MRI and its group company not only receive COVID-19, but also accept thousands of samples for various clinical tests every day. Employing human to handle their test tubes is a nightmare. Under this background, we started considering automatic tube manipulation. Our expectation is that a person puts a rack with mixed and non-arranged tubes in front of a robot. The robot performs recognition, reasoning, planning, manipulation, etc., and returns a rack with separated and arranged tubes. The system should be simple-to-use and there shall be no requests for expert knowledge in robotics.

We develop such a system using a robotic manipulator (Yumi, ABB) and a 3D vision sensor (Phoxi M, Photoneo). We realize a combined task and motion planner to autonomously generate the robot's logical sequences, grasping poses and motion trajectories. Compared with previous research (1, 2, 3), our contribution is the combined planning with backward updates between vision, planning, and execution, which makes the system robust and simple-to-use. Our system workflow is shown in Figure 1. A complete planning-execution-re-exploration demo is available in Movie S1.

We have performed experiments to examine the developed system. In the first experiment, a person drops a rack in the workspace, and the robot arranges the tubes inside the rack. Results show that a person may ask the system to arrange tubes into different sections of the rack by simply specifying new goal patterns (Movie S2). The second experiment is more pragmatic. A robot is requested to move the tubes of a different type to an empty rack. The task is easier since an empty rack is less obstructed. The search gets faster and the planner finds more solutions (Movie S3).

**Figure 1. The workflow of our tube-handling system.** The blue arrows are the main forward stream. The red arrows are the failure stream. The gray arrows denote the data that should be prepared in advance.

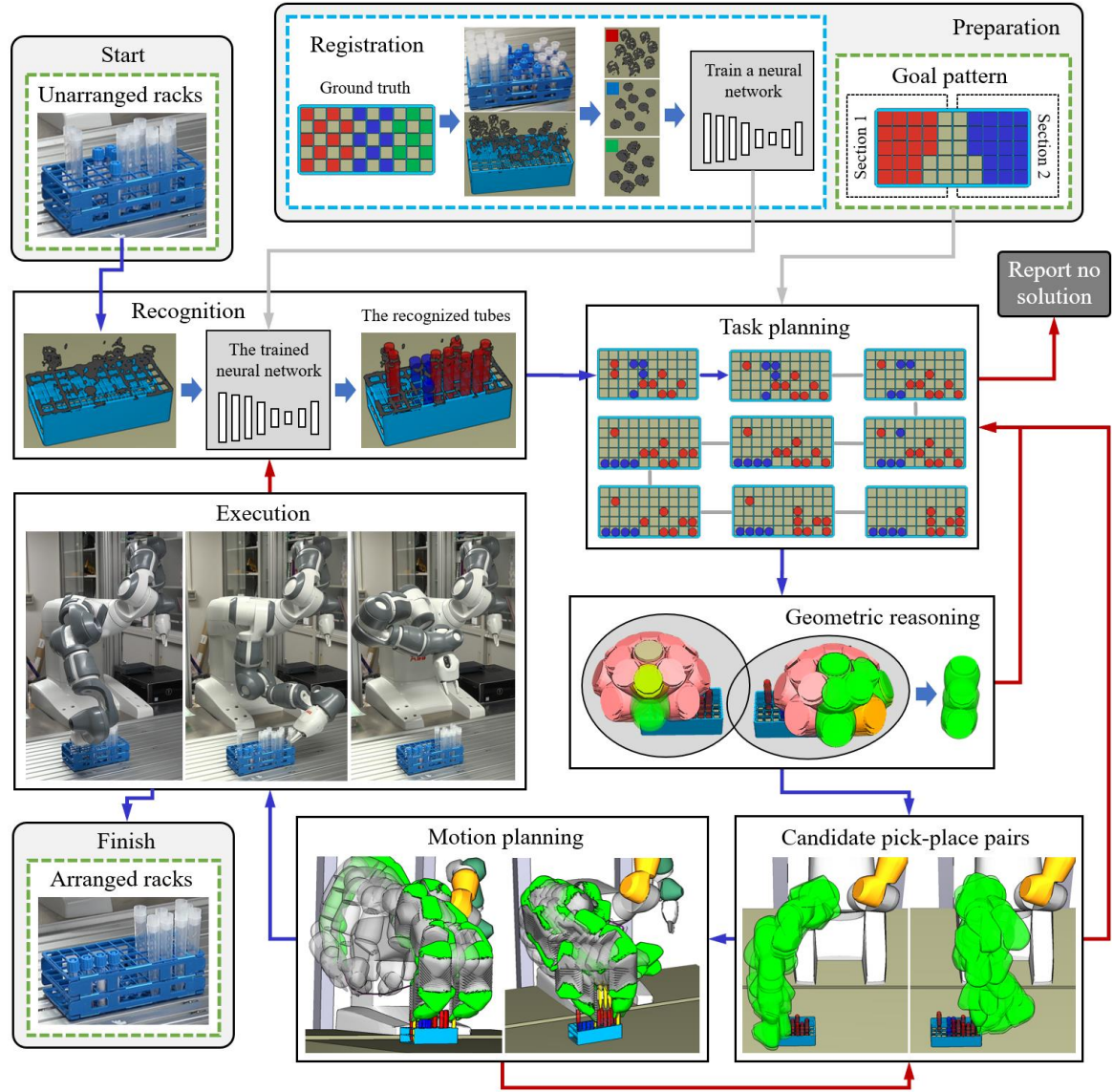

**System workflow**

*Vision*: The first step is recognizing racks and tubes. Since tubes are small and mostly transparent, conventional global search and model-matching based methods (4) are not applicable. Thus, instead of directly detecting tubes, we detect a rack first (5). As tubes are always in the holes of a rack, we can then concentrate on the point clouds inside the rack holes and recognize the tubes locally.

We train a neural network to perform the recognition. A registration process, which is shown inside the "Preparation" box of Figure 1, is needed to collect the training data. During the registration, the tubes are manually arranged following a ground-truth pattern and are dropped under our vision sensor for perception. Since we know the ground truth, we know the correspondence between the perceived point clouds and the types of the tubes, and thus get the data and their labels for training. The trained neural network is later used in the "Recognition" box of Figure 1.

*The combined task and motion planner*: The combined planner has a hierarchical structure. At the highest level, the planner employs adapted heuristic search (6) to find a logical sequence for arrangement. This is illustrated by the "Task planning" box of Figure 1. Here, a goal pattern needs to be prepared in advance to tell the planner where to rest different types of tubes. The goal pattern can be changed following the user's needs. The adapted heuristic search finds a logical sequence of moving the tubes according to the goal pattern. The obtained sequence at this level is logical rather than feasible since there is no consideration of the robotic mechanisms.

In the middle level, the planner employs geometric reasoning (7) to determine if there are available grasps to grab the tubes identified by the logical sequence. Beforehand, we annotate or plan grasps considering the size of the tubes. Then, for a tube at its initial and goal poses in a logical step, we compute their collision-free and IK-feasible grasps. When there is an identical grasping pose for the tube at both its initial and goal poses, we determine the tube can be grasped by the robot. The geometric reasoning process is illustrated in the "Grasp reasoning" box of the figure. It solves geometric conflicts at static poses but there is no consideration about motion.

At the lowest level, the planner employs constrained probabilistic motion planning (8) to find motion trajectories for the robot. Constraints like tube directions are considered during planning to avoid splitting. The trajectories found by the planner in this level are directly executable by a robot.

The lower levels update higher levels and trigger re-explorations recursively to ensure the completeness of the planner. The red arrows in the Figure indicate the flow of this update. A failed grasp reasoning will invalidate the current pick-and-place pair and trigger a new task planning. A failed motion planning will invalidate the current candidate configuration pair and iterate to the next one. It will act in the same way as a failed grasp reasoning if all configuration pairs are invalid.

*Error Recovery*: The planned results are not necessarily successful during execution. The reasons could be wrong visual detection, slippery in the finger pads, changes in rack positions, etc. (9) The system will recover by triggering a new visual detection and restart the combined planner from the "Recognition" box.

**Conclusions**

We have shown that the developed system can arrange test tubes in racks without much human intervention. This owes to the help of the 3D visual detection and the combined planner that plans sequences and trajectories at different levels. We expect such a system to play an important role in helping managing public health (10). In the future, we hope such a system could play more hard-core roles in clinical manipulation like handling mixers and pipettes.

# SUPPLEMENTARY MATERIALS

**Text S1. More explanations about the system input.** There are three inputs to the system: 1. A tube rack with random tubes; 2. The goal pattern; 3. A tube rack with known ground truth for machine learning. Item 1 is needed each time a new task is to be performed. Item 2 is set once when people hope the robot arranges the tubes into a new pattern. Item 3 is required when unforeseen tubes need to be registered. The frequency of these three inputs are 1 >> 2 >> 3. In Figure 1, items 1 and 3 are illustrated by the "Unarranged racks" box and the "Registration" box respectively. They must be the real rack and tubes dropped under the vision sensor since we need to collect point cloud data for them. Especially for item 3, the tubes must be prepared following the ground truth. Item 2 is shown by the "Goal pattern" box. It could either be a real rack dropped by users under the vision sensor or directly inputted into the system by using a keyboard. In the former case, the system detects the goal sections for different tubes using the trained neural network.

**Text S2. More explanations about the detection.** The rack can be found using Iterative Closest Point (ICP)-based template matching since it is made of solid materials and has good geometric features. The workflow is shown in Figure S1. First, we crop the potential point area using segmentation (i.e., DBSCAN, Density-Based Spatial Clustering of Applications with Noise). Then, Principal Component Analysis (PCA) is applied to roughly align the main axes of the cropped point cloud and the template. ICP is used after the alignment for fine adjustment. Compared to the rack, the tubes are difficult to be detected as (1) there are reflective noises that are difficult to tell, (2) they are transparent and the captured point cloud has holes. For these reasons, we train a neural network to automatically learn how to classify them.

**Figure S1. Detecting the rack and extract the tube point cloud clusters.** (a) Sensor configuration. (b) The workflow: Segmentation is done using DBSCAN. Rough pose estimation is done using PCA. Fine pose estimation is done using ICP.

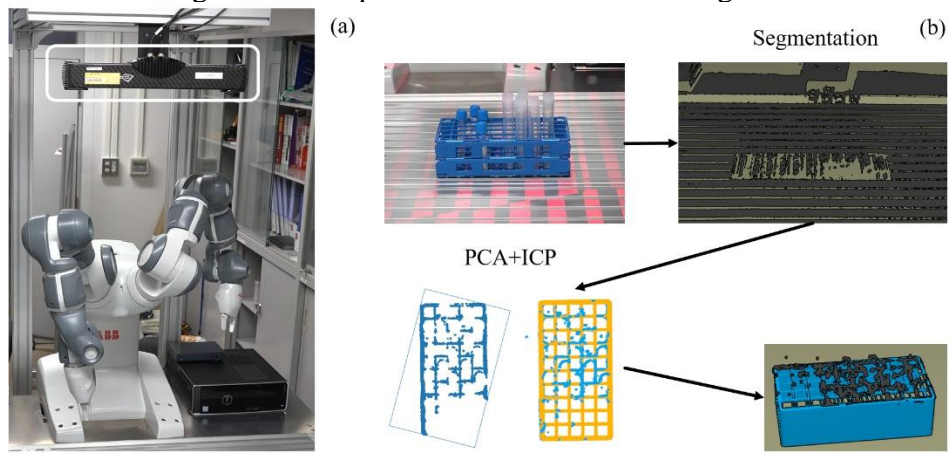

**Text S3. Details of the adapted heuristic search.** In the adapted heuristic search, each state of the tube rack is treated as a node. The cost of expanding into a new node is computed as the total cost of previous nodes plus a heuristic (the displacements of the tubes in the new node from the goal pattern). Whether we can expand to a new node is determined by using simple filters to check whether a tube in one rack hole can be moved to another. The simple filters play the role of a rough collision detector and examine if holes are empty so that gripper fingers or tubes can be potentially inserted. Figure S2(a) shows six exemplary simple filters. These filters roughly assume that a tube can be picked up when its vertical, horizontal, or diagonal adjacent rack holes are empty. Also, it assumes an empty hole could be used to rest tubes under the same condition. The simple filters are a rough estimation of collisions at the logical layer. Exact examinations are done later in geometric reasoning.

Using the filters, the adapted heuristic search expands to new nodes like Figure S2(c). There are several possible children to expand to. The adapted heuristic search determines the next one by selecting the node that has smallest $g(x)+h(x)$, where $g(x)$ is the number of steps to move from the very initial node to the next node (the total cost), $h(x)$ is the number of tubes that are not in the expected sections of the goal pattern (the heuristic). The computation is illustrated in Figure S2(b). When two children have the same $g(x)+h(x)$, the algorithm will select the one that has a smaller $h(x)$.

**Figure S2. Filtering and expansion in the adapted heuristic search.** (a) Using simple filters and a weight map to determine expansion. (b) The heuristic functions. (c) An exemplary expansion.

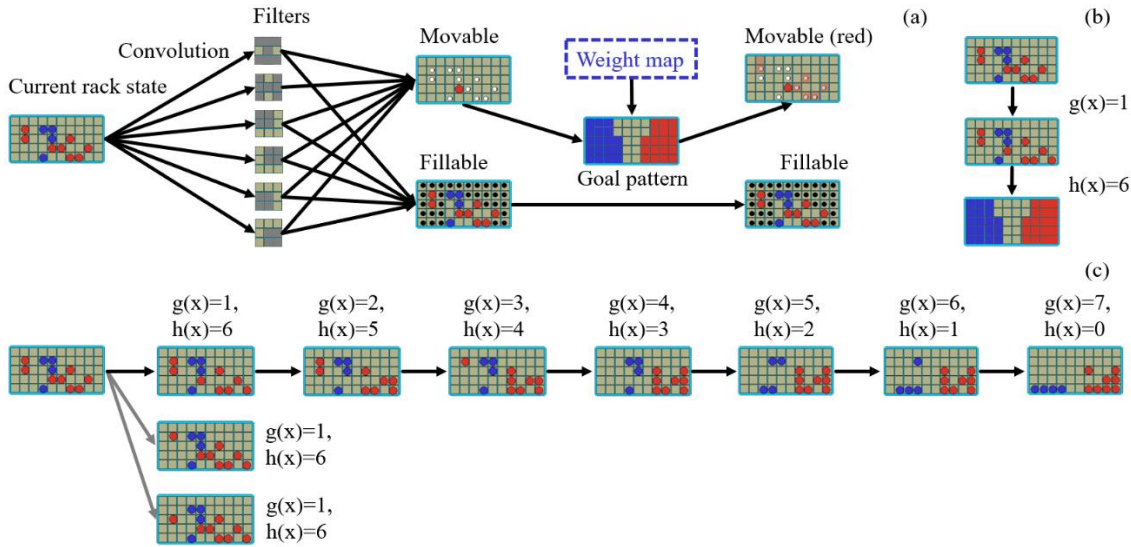

**Text S4. The backward updates.** The backward updates evolve the adapted heuristic search by using a weight map to trace the failures. The weight map is essentially a list of records remembering the failures at the rack holes. When failure happens, the planner adds a record to the list so that the heuristic search could avoid expanding to recorded nodes. The "Weight map" dash box in Figure S2(a) and the red arrows pointing into the "Task planning" box in Figure 1 show where the weight map functions.

**Text S5. Investigating re-explorations and backward updates of the example in the main article.** Figure S3 shows the complete re-explorations and backward updates during planning the task in the main article. The abbreviations TP, GR, and MP represent Task Planning, Grasp Reasoning, and Motion Planning respectively. The numbers after them are the sequential ID. The blue arrows indicate success in both grasp reasoning and motion planning. The orange arrows indicate a successful in grasp reasoning but failed motion planning. The red arrows indicate a failed grasp reasoning (In this case, there are no candidate configuration pairs and the workflow cannot move forward to motion planning.). When there is a failure, either it would be an MP failure or GR failure, the system will update the weight map and trigger a re-exploration. The blue dash arrows in Figure S3 indicate the re-explorations. The example in the figure triggered 7 re-explorations before it finds a solution. The successful logical sequence is highlighted with blue shadows for the reader's convenience. It has 8 logical steps.

Detailed visualization of the failures is illustrated in Figure S4. Here, the bouquets after the logical steps illustrate the grasp reasoning. Each model in the bouquet is a candidate grasping pose. The ones colored by green indicate feasible grasps. The pink ones indicate collided grasps. The orange ones indicate IK-infeasible grasps. Grasp reasoning fails when there are no common green grasps between two nodes. In the same figure, the simulated robots after the bouquets illustrate the motion planning. A red robot arm indicates motion planning failed at that configuration. The time-lapse arms illustrate the successfully planned motion between the green configurations (The green configurations are the IK solutions of the green grasping poses). The abbreviations and their sequential IDs are identical to those in Figure S3. The successful grasp reasoning, motion planning, and their correspondent logical sequence are also highlighted with blue shadows for the reader's convenience. Note that Figure S4 is not complete. Some parts of TP6 and TP7 are omitted to avoid a too large figure size. The final motion found by the planner is shown in Figure S5. Five of the motion pieces (MP1.0, MP6.2, MP7.0, MP7.1, MP7.2) move the red tubes. The remaining (MP3.0, MP6.0, and MP6.1) moves the blue tubes.

**Figure S3. The complete re-explorations and backward updates during planning the task in the main text.** TP, GR, and MP represent Task Planning, Grasp Reasoning, and Motion Planning respectively. A blue arrow indicates success in both grasp reasoning and motion planning. An orange arrow indicates a successful grasp reasoning but failed motion planning. A red arrow indicates a failed grasp reasoning. A blue dash arrow indicates a re-exploration.

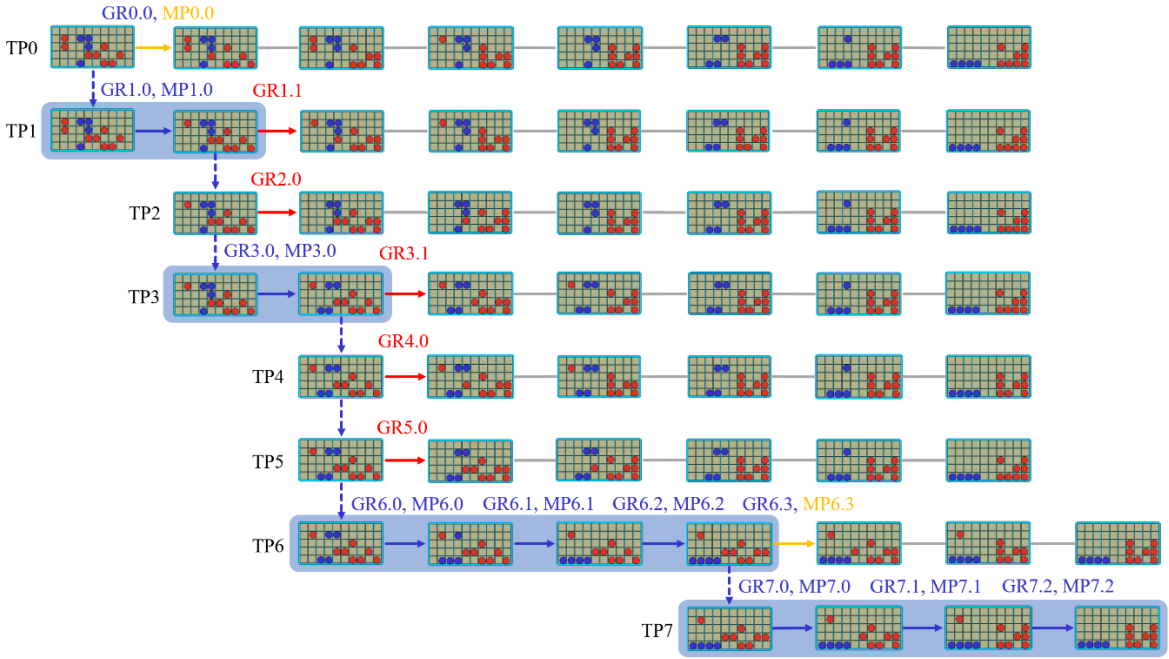

**Figure S4. Detailed visualization of the failures and re-explorations in Figure S3.** The IDs and colors are identical to those in Figure S3. The bouquets after the logical steps illustrate the grasp reasoning. Each model in the bouquet is a candidate grasping pose. Green models indicate collision-free and IK-feasible grasps. The simulated robots after the bouquets illustrate the motion planning. A red robot arm indicates motion planning failed

at that configuration. The time-lapse arms illustrate the successfully planned motion. The figure is not complete. Some parts of TP6 and TP7 are omitted to avoid a too large figure size.

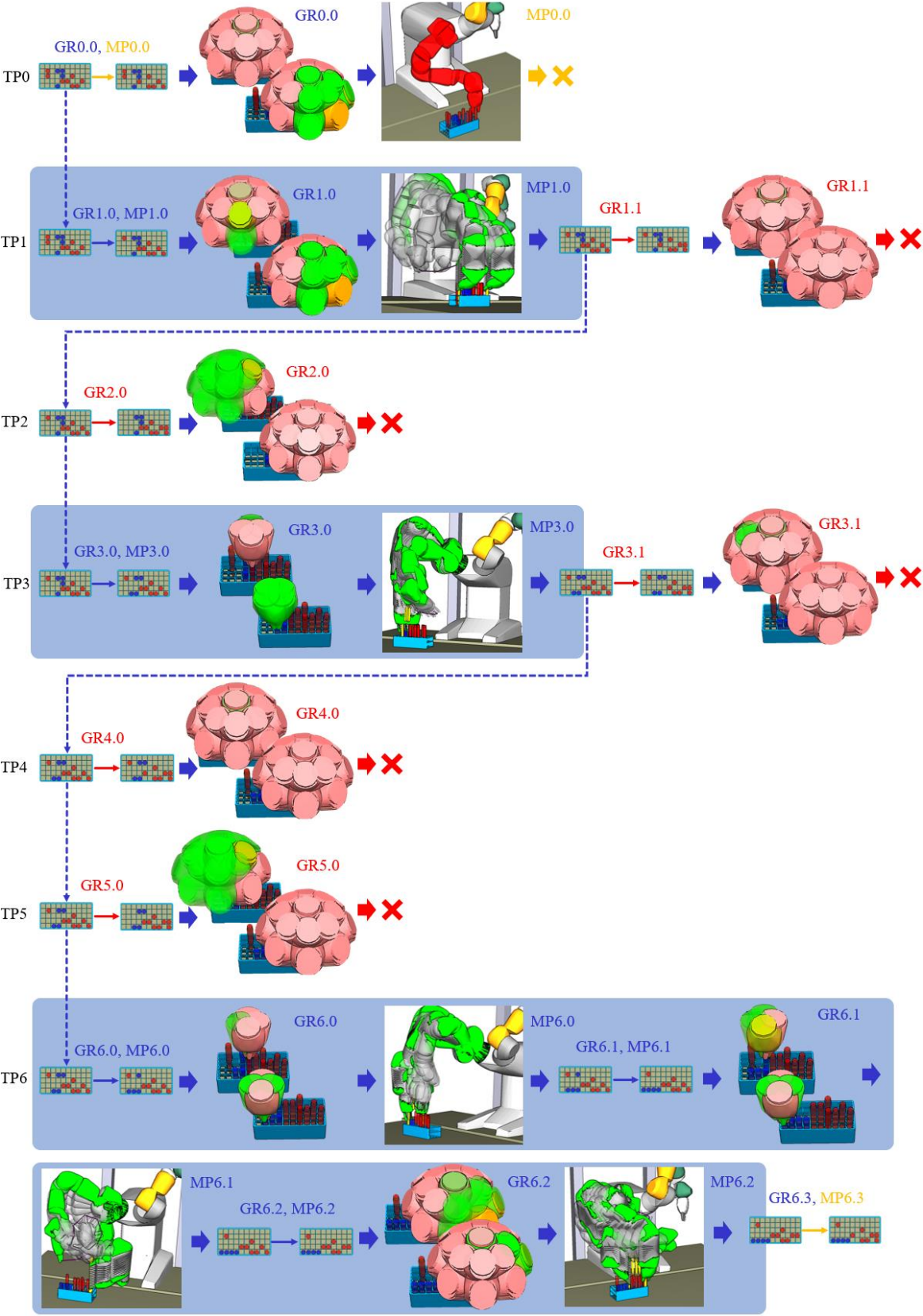

**Figure S5. The final motion trajectories found by the planner.** The motion piece IDs on the upper right corners are identical to those shown in Figure S3 and S4.

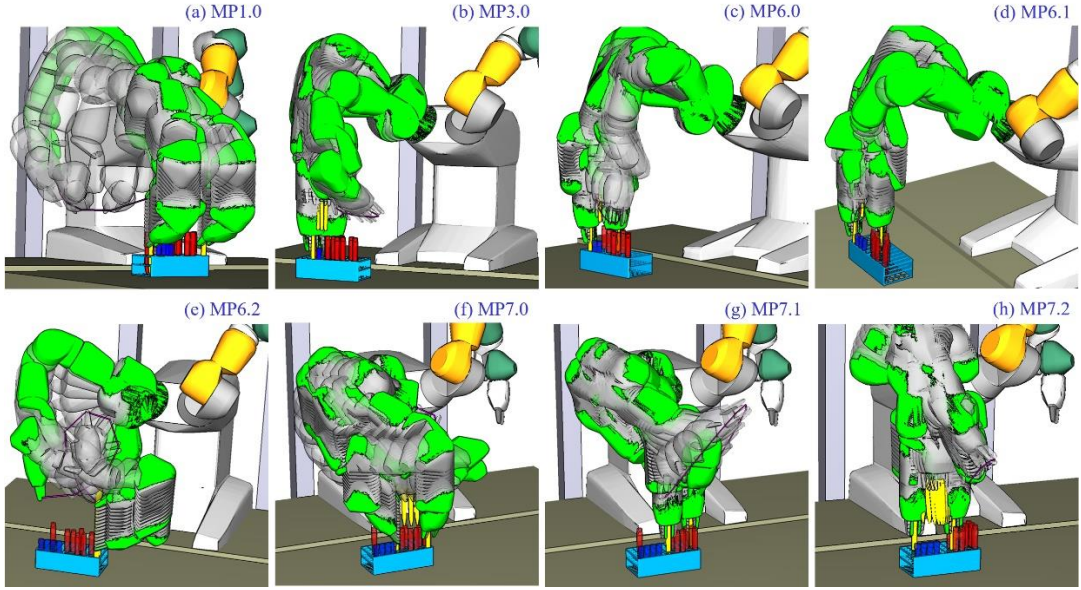

**Text S6. Execution and error recovery.** The real-world execution of the planned motion in Figure S5 is shown in Figure S6. The motion piece IDs in the sub-titles are identical to those in Figure S5. During the execution, the robot encounters a motor supervision error (overload) at (f) MP6.2. The reason is the squeezing force of the robot fingers caused a large reaction force from the table. At the failure, the system restarts by triggering a new visual recognition and combined planning. The robot gets a new sequence shown in TP9 and successfully re-plans and executes the motion in (i)-(l). A video showing the details of the plan-execution flow is in Movie S1. Some other examples could be found in Movie S2.

**Figure S6. Executing the planned trajectories.** The robot gets a motor supervision error (overload) at (f) MP6.2, performs a re-planning in (h), and successfully executes the new motion in (i)-(l).

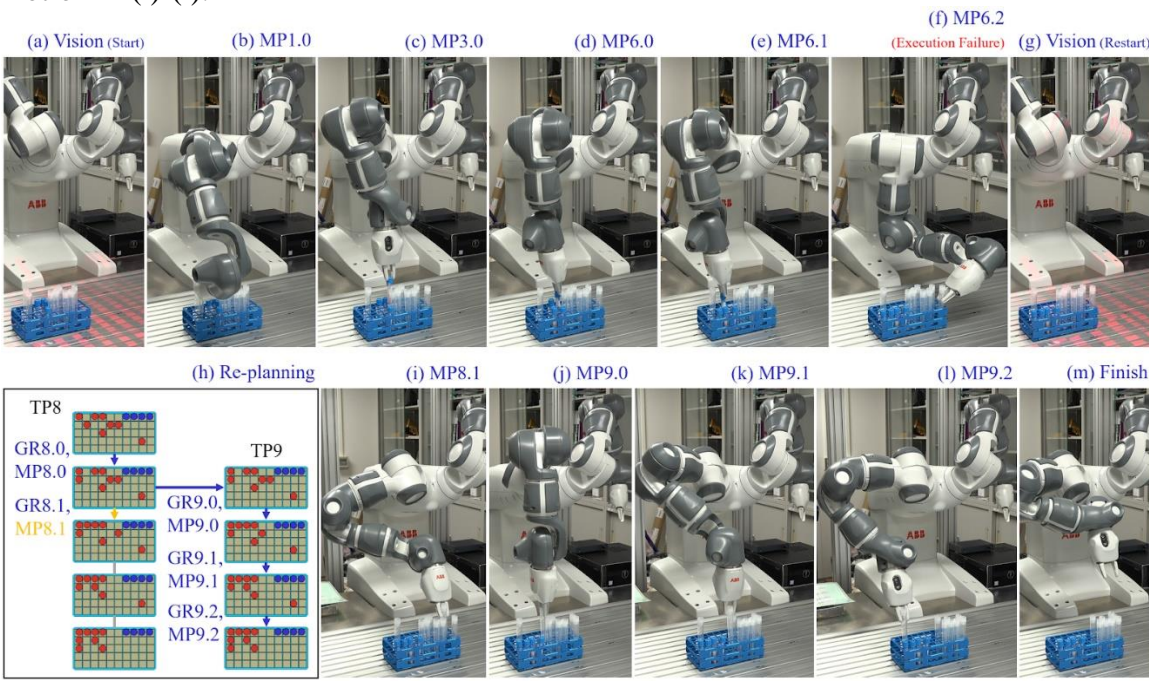

**Movie S1. A complete planning-execution-re-exploration demo.**

**Movie S2. Two other examples.** The robot is asked to arrange the tubes following different goal patterns.

**Text S7. Separating tubes into a new rack.** When the goal is to separate tubes of a different type to a new rack, we do the following modification: 1. The node in the adapted heuristic search is changed to incorporate both states of the original rack and the new rack. 2. The simple filter additionally requires a tube must be picked up from the original rack and placed down into a hole in the new rack. Movie S3 shows two examples of separating into a new rack.

**Movie S3. Using the same framework to plan separating tubes into a new rack.** The robot in the right video raises lower. Thus, the motion planning bears more workload.

**Text S8. The workload of motion planning.** The pick-up and place-down motion are treated as a linear movement along the depth direction of the rack holes. Motion planning plays the role between the end of picking up and the start of placing down. The length of the linear movement is crucial for motion planning. A low length increases the workload of motion planning. On the other hand, a long length reduces the success rate to find a feasible motion. But the workload of motion planning is lighter and the planned trajectory is simpler and preferable. Figure S7 shows the comparison. Details can also be found in Movie S3.

**Figure S7. Comparing the workload of motion planning.** (a.1-4) The tube is raised higher. The planned motion is a straight line. The motion is preferable but the robot is sometimes not able to raise the tube that high. (b.1-4) The tube is raised lower. Motion planning bears a larger workload.

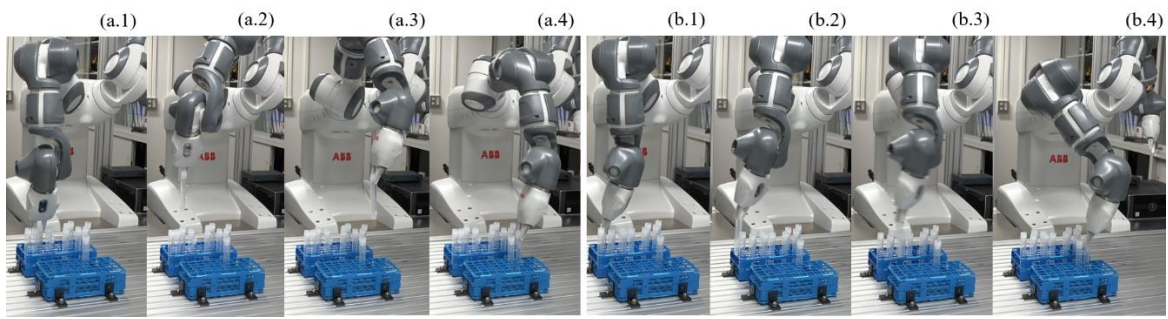

**Author contributions:** W. Wan designed and developed the whole system, as well as led the writing of the paper. T. Kotaka contributed to the conception and financial supports. K. Harada provided suggestions to essential algorithms and contributed to the management.
**Competing interests:** The authors declare that they do not have competing interests.